\documentclass[10pt,twocolumn,letterpaper]{article}
\usepackage[pagenumber]{cvpr}

\usepackage{pifont}
\usepackage{xcolor}

\newcommand{\benchyes}{\textcolor{green!45!black}{\ding{51}}}
\newcommand{\benchno}{\textcolor{red!75!black}{\ding{55}}}

\usepackage[T1]{fontenc}
\usepackage{graphicx}
\usepackage{amsmath,amssymb}
\usepackage{booktabs}
\usepackage{multirow}
\usepackage{array}
\usepackage{xspace}

\graphicspath{{figures/}}

\definecolor{cvprblue}{rgb}{0.21,0.49,0.74}
\usepackage[pagebackref,breaklinks,colorlinks,allcolors=cvprblue]{hyperref}
\def\confName{CVPR}
\def\confYear{2027}
\title{InfoSceneBench: Benchmarking Joint Constraint Satisfaction in Information-Rich Image Generation}

\author{
Li-Ang Xu$^{1}$ \quad
Xian-Ling Mao$^{1}$ \quad
Rong-Cheng Tu$^{2}$ \quad
Zi-Ao Ma$^{1}$ \quad
Zhijing Wu$^{1}$ \\
$^{1}$Beijing Institute of Technology \quad
$^{2}$Nanyang Technological University\
}

\begin{document}
\maketitle

\begin{abstract}
Generating information-rich images---such as posters, flowcharts, and technical diagrams---requires text-to-image models to render specified text accurately, bind visual attributes to the intended elements, follow prescribed layouts, and ensure that labels, values, and other scene elements preserve their intended associations and logical relationships. Existing benchmarks typically assess text rendering separately from attribute binding and layout, while more comprehensive assessments have largely focused on specific domains, such as commercial design. However, whether models can jointly satisfy these requirements across diverse information-rich scenes remains insufficiently evaluated. To address this gap, we introduce \textbf{InfoSceneBench}, a benchmark comprising 1,260 prompts across six scene families, 31 subcategories, and three difficulty levels, with each prompt combining requirements for text, visual attributes, layout, and information relationships. We further equip InfoSceneBench with \textbf{InfoScene-Eval}, a four-dimensional evaluation framework assessing text rendering, attribute binding, layout control, and semantic-functional consistency through OCR-based text matching, targeted visual verification, and independent holistic assessment. Experiments with 16 generation configurations reveal distinct capability profiles across the four dimensions and lower performance at higher difficulty levels. On 385 human-annotated images, InfoScene-Eval achieves a Spearman correlation of 0.856 with human judgments. The benchmark dataset and evaluation code are provided in the supplementary material.
\end{abstract}

\section{Introduction}
\label{sec:intro}

Recent advances in text-to-image generation~\cite{rombach2022high,saharia2022photorealistic} have enabled the creation of increasingly sophisticated visual content, including posters, flowcharts, technical diagrams, and digital interfaces. In these information-rich scenes, text, visual organization, and relationships among scene elements jointly determine how information is conveyed. For example, a receipt must associate prices with the correct products, while a technical diagram must connect labels to their corresponding components. Even when the text is rendered accurately, incorrect associations can change the information conveyed by the image. Generating information-rich images therefore requires models not only to reproduce specified text and follow visual requirements, but also to preserve the intended information and relationships within a coherent scene.

Existing benchmarks have examined different aspects of visual
text generation and instruction following. Early efforts focus on the accuracy of rendered text: STRICT~\cite{zhang2025strict} examines text correctness and legibility under varying text lengths,  while OCRGenBench~\cite{zhang2025ocrgenbench} extends evaluation to text generation, editing, and OCR-related tasks. For multi-region text generation, CVTG-2K~\cite{du2025textcrafter} focuses on accurate and legible rendering of specified text instances, while UltraText Bench~\cite{liu2026ultratext} extends the evaluation to dense bilingual text with spatial and visual requirements. Beyond text-centric evaluation, more comprehensive assessments have emerged in specific visual domains. BizGenEval~\cite{li2026bizgeneval} examines multiple dimensions of visual correctness in commercial design, while IGenBench~\cite{tang2026igenbench} focuses on data and semantic reliability in infographic generation. Meanwhile, evaluation methods such as TIFA~\cite{hu2023tifa} and DSG~\cite{cho2024davidsonian} use prompt-derived
questions to assess fine-grained image--text alignment. Together, these studies provide complementary approaches to textual fidelity, compositional constraints, and domain-specific information correctness.

\begin{figure*}[t]
    \centering
    \includegraphics[width=\textwidth]{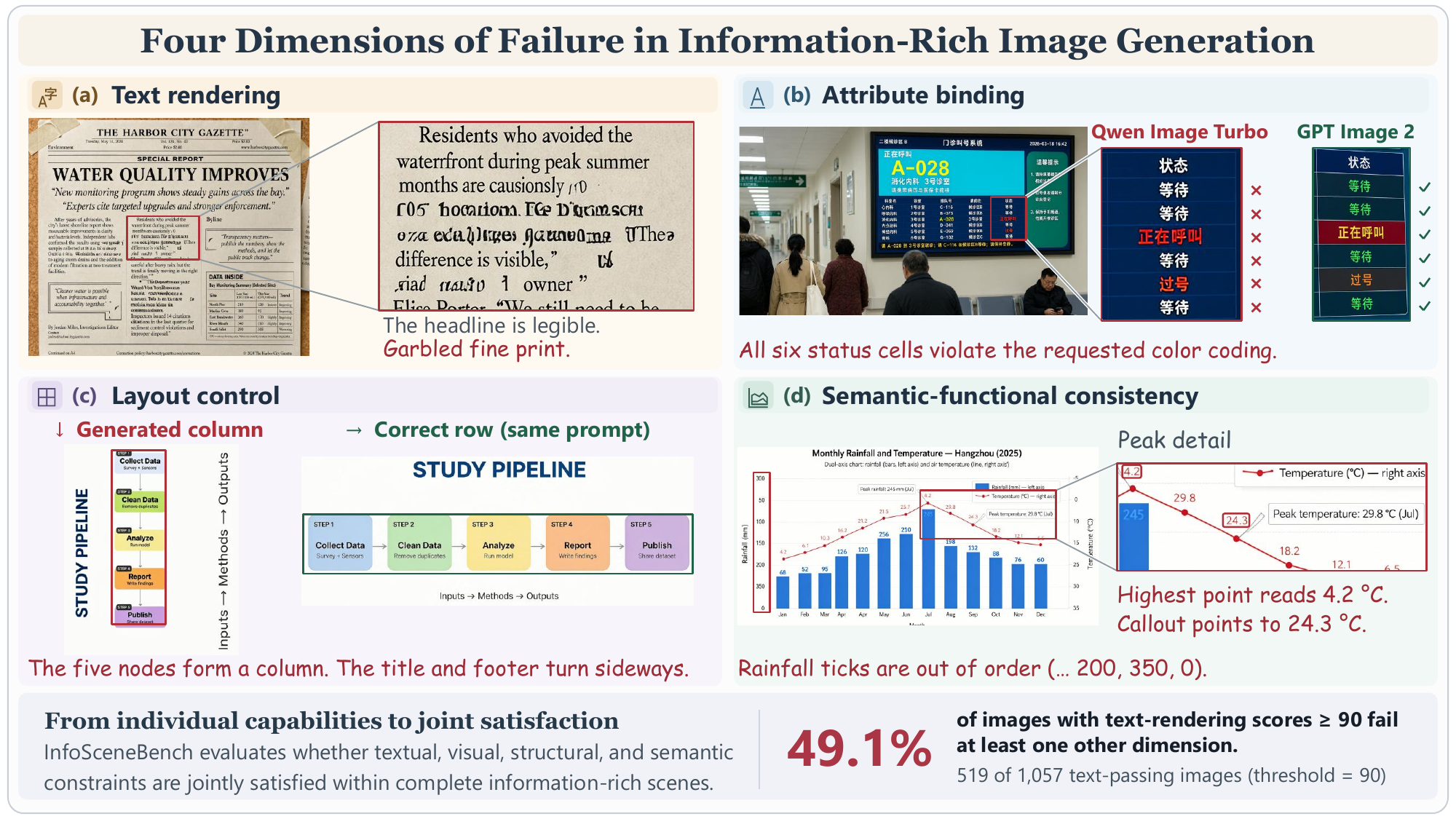}

\caption{
\textbf{Illustrative failures of joint constraint satisfaction in information-rich image generation.}
Four examples from InfoSceneBench:
(a) \textit{Text rendering (T)}: a legible newspaper headline with corrupted body text;
(b) \textit{Attribute binding (A)}: incorrectly colored status labels in a queue display;
(c) \textit{Layout control (L)}: a five-step workflow arranged vertically instead of horizontally;
and (d) \textit{Semantic-functional consistency (F)}: a peak-temperature annotation inconsistent with the indicated data point.
Red highlights mark violations, while green highlights indicate compliant same-prompt examples.
At a threshold of 90, 49.1\% (519/1,057) of text-passing images fail at least one of A, L, or F, showing that text accuracy alone does not ensure joint constraint satisfaction.
}

\label{fig:teaser}

\end{figure*}

Despite these advances, the joint satisfaction of text rendering, attribute binding, layout control, and semantic-functional requirements remains insufficiently evaluated across diverse information-rich scenes. Fig.~\ref{fig:teaser} illustrates four representative failures. In (a), a newspaper headline is legible while the body text is corrupted, revealing a failure in text rendering. In (b), a queue display reproduces the status labels but assigns them incorrect colors, violating the specified attribute bindings. In (c), a five-step workflow is arranged vertically rather than in the required horizontal sequence, failing to follow the specified layout. In (d), a chart's peak-temperature annotation points to a data value inconsistent with its label, illustrating a failure in semantic-functional consistency. Unlike errors in text reproduction or visual organization, such inconsistencies can change the information conveyed by the image, even when individual labels and graphical elements appear correct. Together, these examples show how satisfying some requirements does not guarantee their joint satisfaction, motivating a benchmark for assessing these interdependent constraints across diverse information-rich scenes.

In this work, we investigate \textbf{information-rich image
generation} under jointly specified textual, visual, and
semantic constraints. The difficulty of these tasks extends
beyond text length or density, arising also from multi-region
organization, visual attribute assignments, hierarchical
structures, and dependencies among information-bearing
elements. To systematically examine these challenges,
we introduce \textbf{InfoSceneBench}, comprising 1,260
prompts across six scene families and 31 subcategories.
The benchmark covers diverse information-bearing artifacts,
including reading materials, transactional documents,
technical graphics, physical signage, and digital interfaces.
Rather than treating textual, visual, and semantic requirements
as isolated tests, its prompts are designed to combine
multiple constraints within complete scenes, including
requirements associated with shared information entities
and relationships across visual elements. Prompts are
organized into three design-difficulty levels with
increasing textual, structural, and relational demands.
A manually reviewed subset of 281 prompts is selected
for generation-system evaluation.

To evaluate joint constraint satisfaction, we equip InfoSceneBench with \textbf{InfoScene-Eval}, a four-dimensional evaluation framework covering \textbf{text rendering (T)}, \textbf{attribute binding (A)}, \textbf{layout control (L)}, and \textbf{semantic-functional consistency (F)}. InfoScene-Eval integrates two complementary assessment branches. The evidence-based branch combines OCR-based text matching for T with prompt-specific visual verification questions for A, L, and F, providing targeted assessments of individual constraints. In parallel, an independent holistic branch evaluates the generated image against the complete prompt across the same four dimensions. The two branches are combined through fixed equal-weight fusion, producing dimension-level and overall scores while retaining detailed evaluation records.

We evaluate 16 generation configurations on the manually reviewed subset of 281 prompts. The results reveal distinct capability profiles across evaluation dimensions and scene families, with all configurations scoring lower on D3 than on D1. Notably, at a strict threshold of 90, 49.1\% of images meeting the text-rendering criterion fail at least one of the other three dimensions, revealing a gap between individual capability and joint constraint satisfaction. On 385 human-annotated images, InfoScene-Eval achieves an overall Spearman correlation of \textbf{0.856} with human judgments.

Our main contributions are summarized as follows:

\begin{itemize}
    \item \textbf{InfoSceneBench:} We introduce a benchmark comprising 1,260 prompts across six scene families and 31 subcategories, with each prompt jointly imposing requirements on text rendering, attribute binding, layout control, and semantic-functional consistency.

    \item \textbf{InfoScene-Eval:} We develop a four-dimensional
    evaluation framework combining OCR-based text matching,
    targeted visual verification, and independent holistic
    assessment to provide complementary and inspectable
    measures of generated images.

    \item \textbf{Systematic evaluation and analysis:}
    We evaluate 16 generation configurations, revealing distinct capability profiles, difficulty-dependent performance differences, and substantial gaps between individual and joint constraint satisfaction, while validating the reliability of InfoScene-Eval against human judgments.
\end{itemize}

\section{Related Work}
\label{sec:related}

\paragraph{Visual Text Generation Benchmarks.}
Visual text generation has advanced through methods such as TextDiffuser, TextDiffuser-2, GlyphControl, AnyText, AnyText2, and TextCenGen~\cite{chen2023textdiffuser,chen2024textdiffuser,yang2023glyphcontrol,tuo2024anytext,tuo2024anytext2,liang2024textcengen}.

TextInVision~\cite{fallah2025textinvision} examines textual characteristics and prompt complexity, while STRICT~\cite{zhang2025strict} studies text accuracy and legibility under varying text lengths. OCRGenBench~\cite{zhang2025ocrgenbench} broadens evaluation to text generation, editing, and OCR-related tasks. For multi-region text rendering, CVTG-2K~\cite{du2025textcrafter} emphasizes accurate and legible rendering of multiple specified text instances, while UltraText Bench~\cite{liu2026ultratext} evaluates dense bilingual text with spatial and visual requirements. InfoTextBench~\cite{xiang2026glyphanchor} further examines text-rich generation and editing under challenging rendering conditions. These benchmarks characterize text fidelity and visual presentation under increasingly demanding rendering conditions.

\paragraph{Compositional and Semantic Benchmarks.}
Beyond text rendering, GenEval~\cite{ghosh2023geneval} and T2I-CompBench++~\cite{huang2025t2i} evaluate general compositional capabilities, including object attributes and spatial relationships. Other work studies compositional evaluation, controllable difficulty, and semantic alignment~\cite{huang2023t2i,li2024evaluating,wu2024conceptmix,hu2024ella}. T2LSC-Bench~\cite{wang2026t2lsc} investigates localized semantic control, examining whether rendered text unintentionally alters surrounding visual content. Related benchmarks further test factuality, instruction following, semantic alignment, and reasoning~\cite{huang2024t2i,wei2026tiif,wang2025unigenbench++,sun2026t2i}. More comprehensive evaluations have also emerged for specific types of visual content. BizGenEval~\cite{li2026bizgeneval} assesses text rendering, layout control, attribute binding, and knowledge-based reasoning in commercial visual design, while IGenBench~\cite{tang2026igenbench} examines data and semantic reliability in infographic generation. These studies extend compositional evaluation toward information correctness, although the joint satisfaction of text rendering, attribute binding, layout control, and semantic-functional consistency across diverse scenes remains less systematically examined.

\paragraph{Fine-grained and Holistic Evaluation.}
Existing evaluation methods provide different perspectives on prompt compliance. TIFA~\cite{hu2023tifa} and DSG~\cite{cho2024davidsonian} use prompt-derived questions for fine-grained verification, with DSG further modeling dependencies among individual checks. VPEval~\cite{cho2023visual} employs specialized visual modules, including OCR and visual question answering, while VQAScore~\cite{lin2024evaluating} measures broader image--text alignment. Explainable, task-decomposed, and reasoning-based evaluators have also been explored~\cite{ku2024viescore,tu2025automatic,ma2025t2i}. Question-based diagnosis has additionally been applied to multi-talker audio-video generation~\cite{zhou2026mtavg,li2026mtavg}. Preference-oriented evaluation is represented by ImageReward, HPS, PickScore, and multi-dimensional reward models~\cite{xu2023imagereward,wu2023human,kirstain2023pick,zhang2024learning,xu2026visionreward}. Fine-grained verification enables the diagnosis of individual requirements, whereas holistic assessment captures broader prompt-level compliance. These approaches offer complementary perspectives on evaluation granularity and diagnostic interpretability.

\section{InfoSceneBench}
\label{sec:benchmark}

\subsection{Task Formulation and Design Principles}

We formulate information-rich image generation as producing an image $I=G(p)$ from a prompt $p$, without a reference image, while jointly satisfying requirements for \textbf{text rendering (T)}, \textbf{attribute binding (A)}, \textbf{layout control (L)}, and \textbf{semantic-functional consistency (F)}. Beyond text length or density, task complexity depends on how specified text, visual attributes, and layouts interact with information relationships, including label--component mappings, value--field correspondences, and process dependencies. InfoSceneBench therefore tests whether these requirements can be satisfied together within a complete scene, rather than in separate generation tasks.

\subsection{Human-guided Benchmark Construction}

\begin{figure*}[t]
    \centering
    \includegraphics[width=\textwidth]{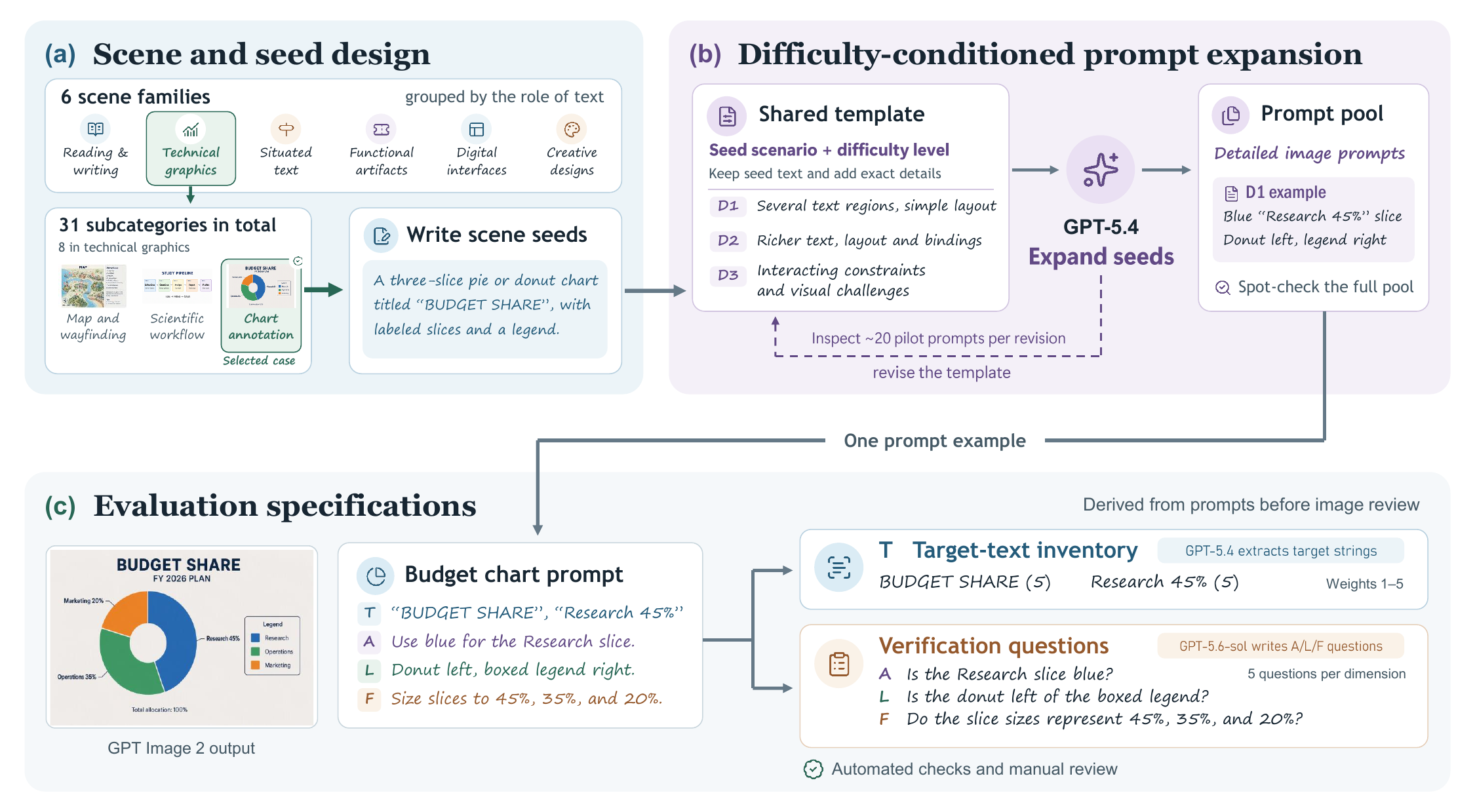}    
\caption{\textbf{Overview of the InfoSceneBench construction pipeline.}
(a) Human-guided scene taxonomy and seed design organize information-rich content into six scene families and 31 subcategories.
(b) Seed scenarios are expanded into detailed prompts across three design-difficulty levels (D1--D3) using a shared template, with iterative pilot inspection and spot-checking.
(c) For the selected evaluation prompts, prompt-specific specifications are constructed, including a weighted target-text inventory for text rendering (T) and five verification questions for each of attribute binding (A), layout control (L), and semantic-functional consistency (F).
The GPT Image 2 output is shown for illustration; evaluation specifications are derived from prompts without examining generated images.
}
    \label{fig:benchmark_construction}
\end{figure*}

As illustrated in Fig.~\ref{fig:benchmark_construction}, we first establish six scene families through a broad review of information-rich visual content and existing benchmarks, further organizing them into 31 subcategories according to the informational roles of text. Within each subcategory, we develop concise seed prompts representing diverse scene instances and their essential information requirements. GPT-5.4 then expands these seeds into detailed generation prompts using a shared instruction template and D1--D3 difficulty guidelines. The template is iteratively refined through pilot inspections of approximately 20 prompts per revision, followed by spot-checking of the full prompt collection.

For generation-system evaluation, we select a 281-prompt subset spanning all subcategories and difficulty levels. Selection also considers diversity in scene instances, textual workload, visual organization, and cross-element relationships. All selected prompts undergo manual review. The final subset comprises 86 D1, 87 D2, and 108 D3 prompts, shared across generation configurations for paired comparisons.

\subsection{Scene Coverage and Constraint Characteristics}

\begin{figure}[t]
    \centering
    \includegraphics[width=0.85\linewidth]{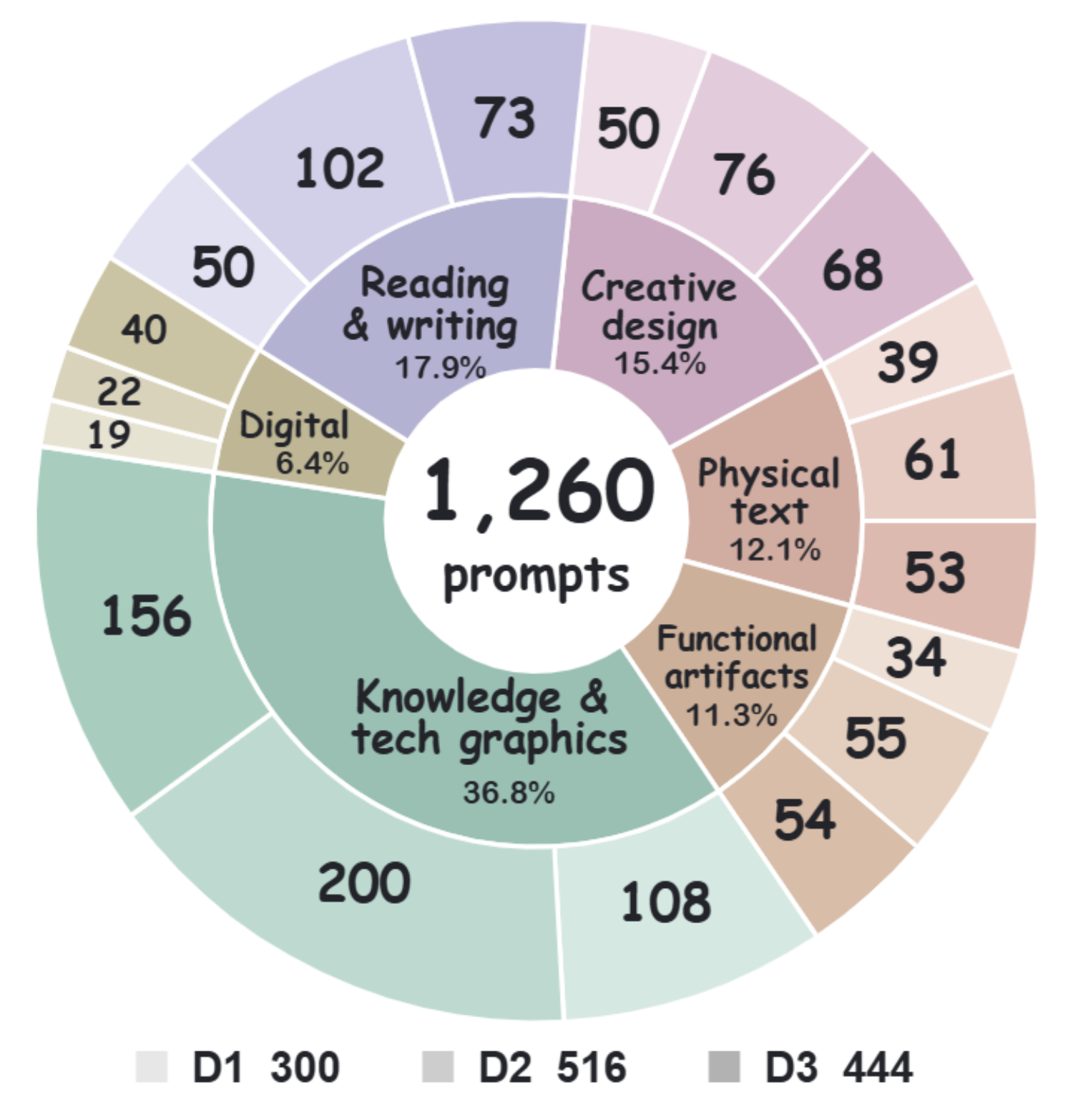}
    \caption{
    \textbf{Scene coverage and design difficulty in InfoSceneBench.}
    The inner ring shows each scene family's share of the 1,260 prompts.
    The outer ring shows prompt counts by difficulty within each family,
    from light (D1) to dark (D3).
    }
    \label{fig:figure3}
\end{figure}

InfoSceneBench contains \textbf{1,260 prompts across six scene families and 31 subcategories} (Fig.~\ref{fig:figure3}): reading and writing media, knowledge and technical graphics, situated physical text, functional and transactional artifacts, digital interfaces and displays, and creative and narrative design. These families cover content ranging from long-form text and annotated graphics to physical signage, structured fields, and interface states. Prompts are organized into D1 (300), D2 (516), and D3 (444). D1 specifies multiple text regions and basic visual organization; D2 introduces more involved layouts, stronger attribute binding, and richer information relationships; and D3 combines more demanding textual, visual, and relational requirements. These are scene-dependent design levels rather than fixed thresholds on text length.

\paragraph{Cross-benchmark task organization and constraints.}
InfoSceneBench differs from existing benchmarks primarily in its joint task design and the prevalence of interdependent requirements. We compare these characteristics with five representative benchmarks through their task organization and a shared rule-based analysis of generation prompts (Table~\ref{tab:constraint_comparison}).

\begin{table*}[t]
\centering
\footnotesize
\setlength{\tabcolsep}{4pt}
\renewcommand{\arraystretch}{1.13}
\resizebox{\textwidth}{!}{%
\begin{tabular}{@{}lcccccc@{}}
\toprule
Metric
& STRICT~\cite{zhang2025strict} & GenEval~\cite{ghosh2023geneval}& CVTG-2K~\cite{du2025textcrafter} & IGenBench~\cite{tang2026igenbench}
& BizGenEval~\cite{li2026bizgeneval} & \textbf{InfoSceneBench} \\
\midrule

Visual focus
& \shortstack{Long-form\\text}
& \shortstack{Object\\composition}
& \shortstack{Multi-region\\visual text}
& \shortstack{Data-rich\\infographics}
& \shortstack{Professional\\visual documents}
& \shortstack{Information-rich\\scenes and artifacts} \\

% Task organization
% & \shortstack{Text-length\\variations}
% & \shortstack{Compositional\\categories}
% & \shortstack{Text regions\\and styling}
% & \shortstack{30 infographic\\types}
% & \shortstack{5 formats $\times$\\4 capability tracks}
% & \shortstack{6 scene families\\3 difficulty levels} \\

Analyzed prompts
& 100 & 553 & 2,000 & 600 & 400
& \textbf{1,260} \\

\midrule
\multicolumn{7}{@{}l}{\textit{Benchmark design}} \\
\addlinespace[2pt]

Joint T/A/L/F task design
& \benchno & \benchno & \benchno & \benchno
& \benchno & \benchyes \\

Dedicated situated physical-text family
& \benchno & \benchno & \benchno & \benchno
& \benchno & \benchyes \\

Image-level T/A/L/F joint test
& \benchno & \benchno & \benchno & \benchno
& \benchno & \benchyes \\

Data--visual consistency tasks
& \benchno & \benchno & \benchno & \benchyes
& \benchyes & \benchyes \\

\midrule
\multicolumn{7}{@{}l}{\textit{Textual workload and constraint density}} \\
\addlinespace[2pt]

Text units / prompt
& 1.0 & 0 & 3.5 & 7.1 & 16.7 & 22.3 \\

Target characters / prompt
& 1,067 & 0 & 39 & 174 & 348 & 267 \\

A / L / F statements / prompt
& 0/0/0 & 0.4/0.2/0 & 1.6/0.2/0
& 0.6/7.2/3.2 & 14.9/30.1/8.1
& 12.1/17.5/4.9 \\

\midrule
\multicolumn{7}{@{}l}{\textit{Constraint co-occurrence and coupling}} \\
\addlinespace[2pt]

Four-dimensional coverage (\%)
& 0 & 0 & 0.2 & 40.3 & 65.8
& \textbf{93.9} \\

Prompts with $\geq$1 text-related relation (\%)
& 0 & 0 & 1.7 & 68.5 & 62.8
& \textbf{83.5} \\

Prompts with $\geq$2 relation types (\%)
& 0 & 0 & 0.1 & 44.3 & 52.5
& \textbf{69.2} \\

Prompts with $\geq$1 coupled text unit (\%)
& 0 & 0 & 0 & 9.7 & 46.8
& \textbf{77.5} \\

\bottomrule
\end{tabular}%
}

\caption{
\textbf{Comparison of benchmark design and constraint characteristics.}
Checkmarks indicate explicit design or evaluation features; crosses do not imply their absence from individual prompts.
Joint T/A/L/F design specifies all four dimensions within each task, while the image-level joint test assesses their simultaneous satisfaction.
A dedicated situated physical-text family denotes a distinct scene category rather than incidental signage.
Quantitative statistics use a shared rule-based prompt analyzer.
Four-dimensional coverage measures detected T/A/L/F co-occurrence, and a coupled text unit is an exact target-text element with detected constraints in all four dimensions.
Text-related relations exclude those identified solely from structured data blocks.
STRICT uses 100 samples constructed from released text resources; other benchmarks use their analyzed prompt collections.
}

\label{tab:constraint_comparison}
\end{table*}

InfoSceneBench is distinguished by the joint organization and coupling of requirements, rather than maximal textual workload. STRICT emphasizes long-form text, with 1,067 target characters per prompt compared with 267 in InfoSceneBench, while GenEval focuses on object composition. CVTG-2K examines multi-region text rendering, but text-related relations are detected in only 1.7\% of its prompts, compared with 83.5\% in InfoSceneBench. More closely related, IGenBench and BizGenEval evaluate information correctness within infographics and professional visual content, respectively. However, four-dimensional requirements are detected in 93.9\% of InfoSceneBench prompts, compared with 40.3\% and 65.8\%, respectively. Moreover, 77.5\% of InfoSceneBench prompts contain at least one text unit subject to all four dimensions, versus 9.7\% and 46.8\%. These results highlight the prevalence of interdependent requirements across diverse information-rich scenes, rather than simply greater text length or constraint density.

\subsection{Evaluation Specifications}

For each evaluation prompt, we derive two types of specifications before examining generated images: a \textbf{target-text inventory} and \textbf{dimension-specific verification questions}. These specifications are shared across generation systems and provide common references for constraint-focused evaluation.

Using GPT-5.4, we extract required visible strings, including titles, labels, numerical fields, captions, and passages, while distinguishing literal text from descriptive generation instructions. Each entry receives an importance weight from 1 to 5 according to its informational role and visual hierarchy. Expected occurrence counts and restrictions on additional text are also recorded, although they are not explicitly incorporated into the current OCR matching score.

For A, L, and F, GPT-5.6-sol generates five prompt-specific verification questions per dimension, each targeting an important, visually verifiable requirement involving scene elements or their relationships. Automated checks validate specification completeness followed by manual review of question relevance, clarity, and visual verifiability for all 281 prompts. The evaluation subset contains \textbf{6,671 target-text entries and 4,215 verification questions} for the evidence-based branch of InfoScene-Eval.

\section{InfoScene-Eval}
\label{sec:evaluation}

\begin{figure}[t]
\centering
\includegraphics[width=\linewidth]{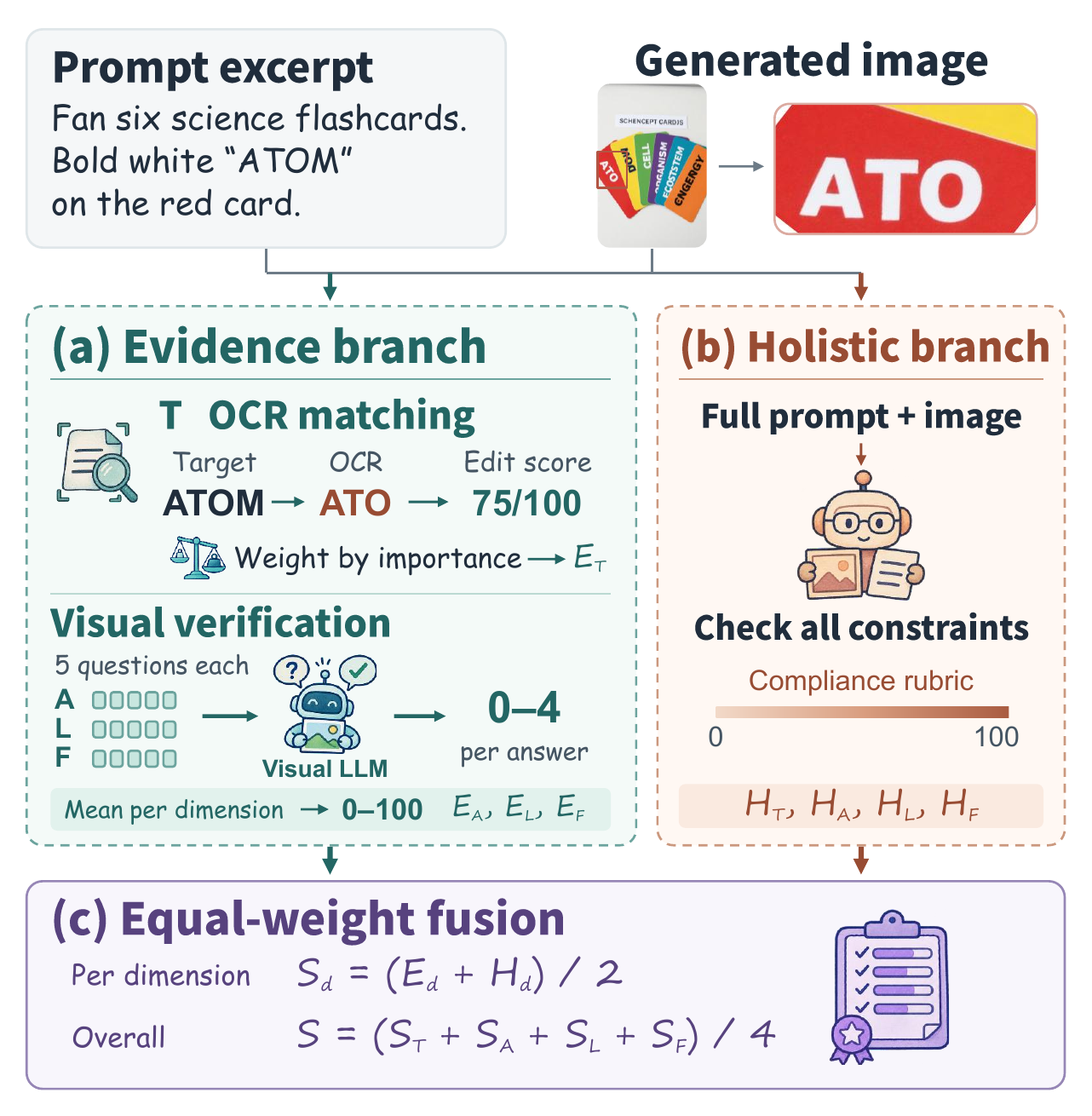}
\caption{
\textbf{Overview of InfoScene-Eval.}
(a) The evidence-based branch combines OCR-based text matching (T) with prompt-specific visual verification (A/L/F).
(b) An independent holistic branch evaluates the complete prompt--image pair across all four dimensions.
(c) Fixed equal-weight fusion produces dimension-level scores, whose mean gives the overall score.
}
\label{fig:pipeline}
\end{figure}

InfoScene-Eval uses two complementary branches (Fig.~\ref{fig:pipeline}). The \textbf{evidence-based branch} combines OCR matching for T with prompt-specific visual questions for A, L, and F. The \textbf{holistic branch} separately assesses the complete prompt--image pair across all four dimensions, including requirements beyond the selected questions. Both MLLM-based assessments use Gemini 3.1 Pro Preview under separate instructions. Fixed equal-weight fusion produces dimension-level and overall scores while retaining interpretable evaluation records.

\subsection{OCR-based Text Evidence}

For text rendering (T), we use \textbf{PaddleOCR 3.7.0}~\cite{cui2025paddleocr} to recognize text in generated images and match it against the prompt-derived target-text inventory (Sec.~3.4). Both target strings and OCR outputs are normalized using Unicode NFKC and whitespace removal. A two-stage matching procedure first aligns short targets with OCR blocks based on similarity, then applies partial matching to longer or unmatched targets, accommodating text spanning multiple blocks while limiting block reuse.

For a normalized target string $a$ and matched OCR text $b$, edit-based similarity is defined as
\begin{equation}
\ell(a,b)=100\max\left(0,1-
\frac{\operatorname{ED}(a,b)}{|a|}\right),
\end{equation}
where $\operatorname{ED}$ denotes Levenshtein edit distance. For partial or multi-block matches, similarities are aggregated over aligned character positions, retaining the highest score for overlapping matches and assigning zero to unmatched positions. This yields a target-level score $r_i$.

The text evidence score is computed as
\begin{equation}
E_T=\frac{\sum_{i=1}^{N}w_i r_i}
{\sum_{i=1}^{N}w_i},
\qquad w_i\in\{1,\ldots,5\},
\end{equation}
where $N$ is the number of target-text entries and $w_i$ is the importance weight assigned during specification construction. Unmatched targets receive zero. The score measures recovery of required text; extra text and deviations from expected occurrence counts are not separately penalized.

\subsection{Dimension-specific Visual Verification}

For A, L, and F, the evidence-based branch uses five prompt-specific verification questions per dimension, prepared in Sec.~3.4. The visual judge receives the generated image, dimension definitions, and corresponding questions, but not the complete generation prompt. A standardized evaluation protocol specifies the scoring criteria and judgment requirements, ensuring consistent assessment across images.

Each question is rated on an ordinal scale $a_{d,j}\in\{0,1,2,3,4\}$, from clear failure to full satisfaction, with explanations and observed visual evidence recorded. Missing required elements receive zero, while genuinely unjudgeable or inapplicable questions are excluded.

Let $\mathcal{V}_d$ denote the set of questions with valid scores. The dimension-level evidence score is
\begin{equation}
E_d=\frac{25}{|\mathcal{V}_d|}
\sum_{j\in\mathcal{V}_d}a_{d,j},
\qquad d\in\{A,L,F\},
\end{equation}
mapping the mean rating to a 0--100 scale. Valid questions are weighted equally. These scores verify selected requirements, not the entire prompt.

\subsection{Holistic Assessment and Score Aggregation}

The holistic branch independently evaluates the complete prompt--image pair without access to OCR results, verification questions, or evidence scores. Following a standardized evaluation rubric defining the four dimensions and their scoring criteria, the judge assigns a score $H_d\in[0,100]$ for each dimension, supported by explanations and visual evidence. The rubric emphasizes compliance with specified requirements rather than overall visual quality.

Evidence-based and holistic scores are combined through fixed equal-weight fusion:
\begin{equation}
S_d=\frac{1}{2}E_d+\frac{1}{2}H_d,
\qquad d\in\{T,A,L,F\}.
\end{equation}
The overall image score is the arithmetic mean of the four fused dimension scores:
\begin{equation}
S=\frac{1}{4}\sum_{d\in\{T,A,L,F\}}S_d.
\end{equation}

Fusion weights are fixed without optimization against human judgments. InfoScene-Eval retains OCR matches, question-level ratings, visual evidence, and holistic judgments to support inspection and error analysis.

\section{Experiments}
\label{sec:experiments}

We evaluate joint constraint satisfaction across generation systems, analyze performance variations across dimensions, scenes, and difficulty levels, and validate InfoScene-Eval against human judgments.

\subsection{Experimental Setup}

We evaluate 16 generation configurations
(Table~\ref{tab:system_performance}) on a shared,
manually reviewed subset of 281 prompts.
Both proprietary/API-only and open-weight
configurations are accessed through hosted APIs.
Each configuration generates one image per prompt
using provider-default inference settings, without
prompt rewriting or explicitly enabled enhancement.
We report the fused T/A/L/F scores from InfoScene-Eval,
with their arithmetic mean as the overall score.

\subsection{Joint Constraint Satisfaction Analysis}
\label{sec:joint_analysis}

Individual dimension scores do not fully capture joint constraint satisfaction. We therefore examine whether all four requirements are satisfied simultaneously in each generated image. For configuration $m$ with $N_m$ evaluated images, the passing rate for dimension $d$ and the joint satisfaction rate (JSR) at threshold $\tau$ are defined as
\begin{equation}
P_{m,d}(\tau)=\frac{1}{N_m}
\sum_{i=1}^{N_m}
\mathbb{I}[S_{m,i,d}\geq\tau],
\end{equation}
and
\begin{equation}
\mathrm{JSR}_m@\tau=
\frac{1}{N_m}
\sum_{i=1}^{N_m}
\mathbb{I}\left[
\min_{d\in\{T,A,L,F\}}
S_{m,i,d}\geq\tau
\right].
\end{equation}

\paragraph{Individual versus joint satisfaction.}
Joint satisfaction is substantially lower than
individual-dimension compliance.
At $\tau=90$, the mean individual-dimension
passing rate is 33.42\%, compared with a JSR
of only 11.26\%.
The gap extends below even the weakest
dimension's passing rate, indicating that
failures across dimensions do not fully overlap.
Notably, GPT Image 2, the strongest configuration
overall, achieves a JSR@90 of only 55.87\%.

\paragraph{Text rendering is not sufficient.}
To quantify failures beyond text accuracy, we define the conditional non-text failure rate as
\begin{equation}
R_m(\tau)=
1-\frac{\mathrm{JSR}_m@\tau}{P_{m,T}(\tau)},
\qquad P_{m,T}(\tau)>0.
\end{equation}
At $\tau=90$, 519 of 1,057 text-passing images (49.1\%) fail at least one of A, L, or F. Even for GPT Image 2, this occurs in 40 of 197 text-passing images (20.3\%). Thus, high text accuracy can coexist with incorrect attributes, layouts, or information relationships.

\subsection{Performance Across Dimensions, Scenes, and Difficulty}
\label{sec:system_performance}

\begin{table*}[t]
\centering
\footnotesize
\setlength{\tabcolsep}{2.5pt}
\renewcommand{\arraystretch}{1.08}
\begin{tabular*}{\textwidth}{@{\extracolsep{\fill}}ll*{14}{c}@{}}
\toprule
& & \multicolumn{5}{c}{Evaluation scores}
& \multicolumn{6}{c}{Scene family}
& \multicolumn{3}{c}{Design difficulty} \\
\cmidrule(lr){3-7}
\cmidrule(lr){8-13}
\cmidrule(l){14-16}
Rank & Configuration
& T & A & L & F & Overall
& R/W & K/T & Phys. & Func. & Dig. & Creat.
& D1 & D2 & D3 \\
\multicolumn{2}{@{}l}{\textit{Benchmark prompts (281 total)}}
& \multicolumn{5}{c}{}
& 40 & 80 & 36 & 61 & 25 & 39
& 86 & 87 & 108 \\
\midrule
\multicolumn{16}{@{}l}{\textit{Proprietary / API-only configurations}} \\
1 & GPT Image 2
& 89.98 & 94.84 & 93.78 & 95.86 & 93.61
& 92.88 & 94.20 & 93.19 & 93.50 & 93.28 & 93.92
& 94.63 & 94.06 & 92.44 \\
2 & Qwen Image 3.0
& 87.45 & 93.92 & 92.33 & 94.36 & 92.02
& 90.13 & 91.89 & 92.89 & 92.28 & 90.95 & 93.68
& 96.16 & 93.10 & 87.84 \\
3 & Seedream 5.0 Pro
& 84.21 & 93.18 & 92.62 & 92.76 & 90.69
& 88.67 & 91.30 & 91.72 & 90.77 & 87.37 & 92.55
& 92.62 & 92.78 & 87.48 \\
5 & Nano Banana 2.1
& 83.96 & 88.57 & 88.77 & 91.75 & 88.26
& 91.67 & 90.31 & 91.24 & 85.29 & 84.68 & 84.76
& 91.20 & 88.79 & 85.50 \\
6 & GPT Image 2.5 Flare
& 83.18 & 81.92 & 82.40 & 85.02 & 83.13
& 82.49 & 84.26 & 82.38 & 83.00 & 83.41 & 82.20
& 83.88 & 83.03 & 82.63 \\
7 & MAI-Image 2.6
& 78.01 & 84.80 & 83.59 & 84.87 & 82.82
& 82.15 & 83.60 & 83.13 & 83.69 & 81.79 & 80.94
& 86.43 & 84.73 & 78.41 \\
9 & Gemini 3 Pro Image
& 81.48 & 81.49 & 80.44 & 85.20 & 82.15
& 81.21 & 82.88 & 79.69 & 80.32 & 87.65 & 83.25
& 83.36 & 82.49 & 80.92 \\
10 & Seedream 4.5
& 69.64 & 80.44 & 75.16 & 73.29 & 74.63
& 66.07 & 73.89 & 82.35 & 79.45 & 65.01 & 76.43
& 86.39 & 81.13 & 60.03 \\
11 & Grok Imagine Quality
& 64.90 & 70.19 & 70.37 & 71.97 & 69.36
& 65.08 & 66.36 & 72.65 & 74.98 & 65.84 & 70.31
& 79.22 & 75.04 & 56.93 \\
12 & GPT Image 1 Mini
& 59.52 & 69.92 & 69.78 & 70.77 & 67.50
& 65.83 & 67.56 & 68.69 & 68.20 & 64.91 & 68.55
& 77.14 & 69.13 & 58.51 \\
13 & Imagen 4 Ultra
& 55.80 & 65.72 & 65.27 & 64.97 & 62.94
& 57.68 & 58.86 & 67.25 & 71.98 & 61.95 & 59.21
& 79.65 & 67.32 & 46.10 \\
15 & FLUX.2 Pro
& 45.75 & 63.96 & 59.75 & 52.81 & 55.57
& 52.00 & 54.29 & 61.13 & 58.43 & 46.95 & 57.74
& 71.38 & 54.22 & 44.06 \\
\midrule
\multicolumn{16}{@{}l}{\textit{Open-weight configurations (hosted inference)}} \\
4 & Qwen-Image-2.1-Turbo
& 86.68 & 88.44 & 87.12 & 91.34 & 88.39
& 86.91 & 88.86 & 85.51 & 90.39 & 86.90 & 89.45
& 92.29 & 89.98 & 84.02 \\
8 & FLUX.3 Klein Image Base
& 70.42 & 88.21 & 87.44 & 84.12 & 82.55
& 79.32 & 81.46 & 84.76 & 84.35 & 80.17 & 84.77
& 88.45 & 83.44 & 77.13 \\
14 & Z-Image Turbo
& 50.94 & 67.55 & 60.37 & 56.20 & 58.76
& 53.64 & 54.17 & 67.26 & 66.71 & 47.44 & 60.43
& 77.67 & 64.63 & 38.97 \\
16 & Ideogram 4
& 38.72 & 58.65 & 32.41 & 42.28 & 43.01
& 40.99 & 37.09 & 47.81 & 48.02 & 43.08 & 44.94
& 51.44 & 41.58 & 37.46 \\
\bottomrule
\end{tabular*}
\caption{\textbf{Generation-system performance across dimensions, scenes, and difficulty.}
Results for the 16 configurations listed on the 281-prompt subset.
T, A, L, and F denote text rendering, attribute binding, layout control, and semantic-functional consistency.
Overall is the mean fused score; scene and difficulty columns give mean overall scores.
R/W, K/T, Phys., Func., Dig., and Creat. denote reading/writing media, knowledge/technical graphics, situated physical text, functional/transactional artifacts, digital interfaces/displays, and creative/narrative design.
Ranks are global across configurations.}
\label{tab:system_performance}
\end{table*}

\paragraph{Overall performance and capability profiles.}
Generation systems exhibit distinct capability profiles, which can be obscured by similar overall scores. As shown in Table~\ref{tab:system_performance}, overall scores range from 43.01 to 93.61, with GPT Image 2 ranking first, followed by Qwen Image 3.0 and Seedream 5.0 Pro. Text rendering is the lowest-scoring dimension for 13 of the 16 configurations, indicating a common challenge in reproducing specified content. However, the limiting dimension varies across systems. Yet similar overall scores can mask distinct
capability profiles: GPT Image 2.5 Flare
outperforms MAI-Image 2.6 in T, whereas
MAI-Image 2.6 performs better in A and L. Such differences demonstrate that aggregate scores alone cannot fully characterize how models satisfy the benchmark's different requirements.

\paragraph{Scene-dependent capabilities.}
Different information-rich scenes expose distinct relative strengths across generation systems. Despite nearly identical overall scores, Qwen-Image-2.1-Turbo performs better than Nano Banana 2.1 on functional and transactional artifacts (90.39 versus 85.29), whereas Nano Banana 2.1 leads on reading and writing media (91.67 versus 86.91) and situated physical text. These scene families place different demands on structured information, extended textual content, and relationships between text and physical objects. The contrasting performance patterns demonstrate the value of evaluating models across diverse informational settings rather than relying on a single scene category.

\paragraph{Effects of design difficulty.}
Increasing design difficulty consistently reduces generation performance, although models differ substantially in their sensitivity. All 16 configurations score lower on D3 than on D1, with 15 declining monotonically across the three levels. The D1--D3 decrease ranges from 1.25 points for GPT Image 2.5 Flare to 38.70 points for Z-Image Turbo. Relative performance can also change: MAI-Image 2.6 leads GPT Image 2.5 Flare on D1 but trails it on D3. These findings suggest varying robustness to more demanding combinations of textual, visual, and relational requirements, rather than a uniform increase in difficulty across systems.

\subsection{Validation of InfoScene-Eval}
\label{sec:human_validation}

Seventeen annotators evaluated 385 generated
images spanning 35 prompts and 11 systems,
with three ratings collected per image.
After annotation quality control, 1,032
ratings were retained.
Human reference scores are computed by
averaging valid ratings for each image.
InfoScene-Eval achieves strong agreement with
human judgments (Spearman $\rho=0.856$;
Pearson $r=0.886$).
Annotation procedures and quality-control
details are provided in the supplementary material.

\section{Conclusion}

We introduced \textbf{InfoSceneBench} to evaluate
joint constraint satisfaction across diverse
information-rich scenes, together with
\textbf{InfoScene-Eval}, which combines OCR-based
text matching, targeted visual verification,
and holistic assessment.
Experiments across 16 generation configurations
reveal distinct capability profiles and
substantial performance degradation as task
difficulty increases.
Crucially, accurate text rendering alone does not
guarantee satisfaction of attribute binding,
layout control, and semantic-functional consistency.
These findings highlight the importance of
evaluating whether textual, visual, structural,
and semantic-functional requirements are jointly
satisfied in information-rich image generation.

\section*{LLM Usage Statement}
We used ChatGPT and Codex to assist with implementing author-specified code functionality and polishing manuscript language. GPT-5.4 and GPT-5.6-sol assisted in expanding and refining the benchmark dataset, as described in the paper. For the experimental evaluation, the test image set was generated using 16 different text-to-image models. Gemini-3.1-Pro-Preview was employed as the multimodal evaluator within our evaluation framework. The core research ideas, methodological design, experimental setup, and conclusions were independently developed by the authors. All AI-assisted outputs were reviewed and verified by the authors, who take full responsibility for the submitted work.

{\small
\bibliographystyle{ieeenat_fullname}
\bibliography{main}
}
\end{document}